\documentclass{article}
\usepackage{spconf,amsmath,graphicx}
\usepackage{hyperref} 
\usepackage{url} 
\usepackage{booktabs}
\usepackage{amsfonts}
\usepackage{nicefrac}
\usepackage{microtype}
\usepackage{lipsum}
\usepackage{fancyhdr} 
\usepackage{graphicx}
\usepackage{algorithmic}
\usepackage{multirow}
\usepackage{amsmath}
\usepackage{listings}
\usepackage{verbatim}
\usepackage{float}
\usepackage{booktabs}
\usepackage[capitalize]{cleveref}
\crefname{section}{Sec.}{Secs.}
\Crefname{section}{Section}{Sections}
\Crefname{table}{Table}{Tables}
\crefname{table}{Tab.}{Tabs.}

\title{
Cluster Contrast for Unsupervised Visual Representation Learning
}

\name{Nikolaos Giakoumoglou and Tania Stathaki}
\address{Department of Electrical and Electronic Engineering, 
         Imperial College London, 
         London, UK\\
         \texttt{\{nikos, t.stathaki\}@imperial.ac.uk}
}

\begin{document}


\maketitle


\begin{abstract}
We introduce Cluster Contrast (CueCo), a novel approach to unsupervised visual representation learning that effectively combines the strengths of contrastive learning and clustering methods. Inspired by recent advancements, CueCo is designed to simultaneously scatter and align feature representations within the feature space. This method utilizes two neural networks, a query and a key, where the key network is updated through a slow-moving average of the query outputs. CueCo employs a contrastive loss to push dissimilar features apart, enhancing inter-class separation, and a clustering objective to pull together features of the same cluster, promoting intra-class compactness. Our method achieves 91.40\% top-1 classification accuracy on CIFAR-10, 68.56\% on CIFAR-100, and 78.65\% on ImageNet-100 using linear evaluation with a ResNet-18 backbone. By integrating contrastive learning with clustering, CueCo sets a new direction for advancing unsupervised visual representation learning.
\end{abstract}
%


\section{Introduction}

Self-supervised learning focuses on extracting features without relying on manual annotations, increasingly closing the performance gap with supervised training techniques in computer vision. Recent state-of-the-art methods in contrastive learning treat each image and its variations as distinct classes, yielding representations that can discriminate between different images while achieving invariance to image transformations. Key to these methods are two fundamental components: (a) a contrastive loss mechanism \cite{hadsell2006dimensionality} and (b) a series of image transformations \cite{chen2020simclr}. Both elements are crucial for the performance of the resulting models \cite{chen2020simclr, misra2019pirl}. 

Unlike contrastive learning, which primarily focuses on intra-image invariance, clustering-based methods enable exploration of similarities between different images \cite{caron2020swav}. While initial assessments of these methods were generally conducted on smaller datasets \cite{xie2016unsupervised}, recent approaches have advanced clustering-based representation learning to larger scales \cite{caron2020swav}. Specifically, these methods generate pseudo-labels through clustering, which then serve as supervision for subsequent training \cite{caron2019deepcluster}. Recent innovations have simplified this process by performing clustering and network updates simultaneously, enhancing the adaptation of the model to evolving data characteristics \cite{pang2022smog}.

\begin{figure*}[htbp]
    \centering
    \includegraphics[width=1.0\textwidth]{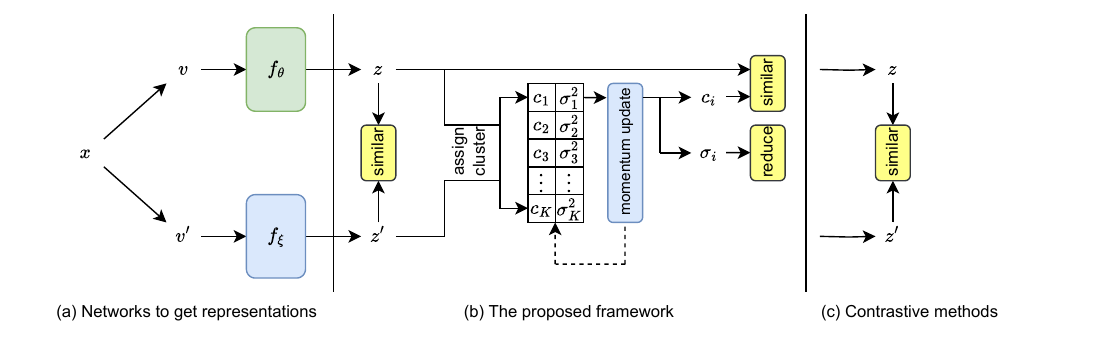}
    \caption{CueCo framework. Like typical contrastive methods, CueCo processes two augmented views, $v$ and $v'$, of the same image $x$ through encoders \(f_\theta\) and \(f_\xi\), producing feature vectors \(\mathbf{z} = f_\theta(v)\) and \(\mathbf{z'} = f_\xi(v')\). In addition to that, CueCo: (a) enforces inter-class separation via contrastive loss, (b) improves intra-class cohesion through clustering, and (c) integrates both for refined feature representation.}
    \label{fig:cueco}
\end{figure*}

In this work, we introduce Cluster Contrast (CueCo), a novel self-supervised learning framework designed to enhance visual representation learning by leveraging the synergistic effects of contrastive learning and clustering methodologies (see \Cref{fig:cueco}). Inspired by the ``push-pull'' dynamics observed in physical forces, CueCo utilizes a dual mechanism that optimizes the feature space: the contrastive component, leveraging the InfoNCE loss \cite{oord2019cpc}, effectively ``pushes'' dissimilar representations apart, akin to repulsive forces, ensuring that distinct classes are well-separated. Concurrently, the clustering component ``pulls'' similar representations closer, analogous to attractive forces, promoting tighter clusters of like data points. 

We make the following \textbf{contributions}: \textbf{(a)} We introduce a novel unsupervised method that employs a dual mechanism to optimize the feature space through ``push-pull'' dynamics, effectively leveraging both contrastive and feature space clustering losses to refine feature discriminability; \textbf{(b)} We provide a comprehensive theoretical analysis of the objectives of our framework, illustrating how the integration of contrastive loss with feature space clustering losses cohesively enhances feature representation; \textbf{(c)} We demonstrate state-of-the-art performance on CIFAR-10, CIFAR-100, and ImageNet-100, underscoring the effectiveness of CueCo.


\section{Related Work} \label{relatedwork}

\subsection{Contrastive Methods}

Contrastive learning employs instance discrimination, where each image is treated as a unique class. This approach pulls together anchor and ``positive'' samples while pushing apart ``negative'' samples \cite{chen2020simclr}. Positive pairs are generated through multiple views like augmentations \cite{chen2020simclr, he2020moco}, patches \cite{oord2019cpc}, or teacher-student models \cite{grill2020byol, caron2021dino}. Training typically uses InfoNCE-based objectives \cite{oord2019cpc, chen2020simclr}. To address computational constraints, methods like MoCo \cite{he2020moco} introduce memory structures. Recent approaches eliminate negative samples \cite{grill2020byol, chen2020simsiam} or use regularization \cite{zbontar2021barlowtwins, bardes2022vicreg,mitrovic2020relic} to prevent collapse, while others enhance learning through synthetic negatives \cite{giakoumoglou2024synco}.

\subsection{Clustering Methods}

Clustering approaches like DeepCluster \cite{caron2019deepcluster} use K-means to form groups and learn features iteratively. While ODC \cite{zhan2020odc} and CoKe \cite{qian2022coke} attempt to improve this process, the inherent delay in supervisory signals remains problematic. SwAV \cite{caron2020swav} reframes grouping as pseudo-labelling with optimal transport, though their equipartition constraints may reduce grouping effectiveness. Alternative approaches include prototype-based methods \cite{li2021pcl}, momentum clusters \cite{pang2022smog}, and contrastive clustering \cite{li2020contrastiveclustering} which treats rows and columns of feature matrices as instance and cluster representations directly. However, our approach differs by introducing separate instance-level and cluster-level queues with both soft and hard assignments, providing more flexible feature learning.


\section{Methodology} \label{methodology}


\subsection{Contrastive Learning} \label{contrastivelearning}

Contrastive learning serves as the backbone of our framework, primarily focusing on scattering features across the feature space to enhance their discriminability. Contrastive learning seeks to differentiate between similar and dissimilar data pairs. We utilize a dual-crop strategy, similar to the methods in \cite{chen2020simclr, he2020moco}, with random data augmentations. Given an image \( x \), and two distributions of image augmentation \( \mathcal{T} \) and \( \mathcal{T}' \), we create two augmented views of the same image using the transformation \(t \sim \mathcal{T} \) and \(t' \sim \mathcal{T}' \), i.e., \(v = t(x)\) and \(v' = t'(x)\). Two encoders, \(f_\theta\) and \(f_\xi\), generate the vectors \(\mathbf{z}=f_\theta(v)\) and \(\mathbf{z'}=f_\xi(v')\) respectively. The learning objective minimizes a contrastive loss using the InfoNCE criterion \cite{oord2019cpc}:

\begin{equation}\label{eq:loss_contrastive}
\mathcal{L}_\text{1}(\mathbf{z}, \mathbf{z'}) = -\log \frac{\exp(\mathbf{z}^\top \cdot \mathbf{z}' / \tau )}{\exp(\mathbf{z}^\top \cdot \mathbf{z}' / \tau) + \sum\limits_{k=1}^{K} \exp(\mathbf{z}^\top \cdot \mathbf{z}_k / \tau)}
\end{equation}

\noindent where, \(\mathbf{z}'\) is \(f_\xi\)'s output from the same augmented image as \(\mathbf{z}\), and \(\{\mathbf{z}_k\}_{k=1}^{K}\) includes outputs from different images, representing negative samples. The temperature parameter \(\tau \) adjusts scaling for the \(\ell_2\)-normalized vectors \(\mathbf{z}\) and \(\mathbf{z}'\). This loss function scatters the feature representations by forcing them to occupy distinct points in the feature space, reducing the likelihood of different instances collapsing into indistinguishable points. Following \cite{he2020moco, chen2020mocov2}, we maintain a queue of negative keys, refreshing only the queries and positive keys in each training batch. A momentum encoder updates the base encoder to ensure consistent representations across updates $\xi \gets m \cdot \xi + (1-m) \cdot \theta$, where $m$ is the momentum coefficient \cite{he2020moco, grill2020byol}.

\subsection{Feature Space Clustering} \label{clustering}

Beyond contrastive learning, our goal is to develop an encoder that groups images of the same class into distinct regions of the feature space. While contrastive learning spreads the feature vectors of different classes, we introduce an objective to pull together features of the same class, forming tight clusters. Since true class labels are not available during pretraining, we apply a clustering algorithm, such as K-means, to infer pseudo-classes. We then gather pre-softmax features, termed activation vectors, from images classified under the same pseudo-class. With this, we can store a running average class prototype, or mean activation vector, $\mathbf{c_i}$, given a key $\mathbf{z'} = f_\xi(x)$ that belongs to the set $S_i$ of features assigned to $i$-th cluster, along the per-class variance \(\sigma_i^2\) via sum of squares as follows:

\begin{equation}\label{eq:centroid}
\mathbf{c}_i = \frac{1}{|S_i|} \sum_{\mathbf{z}' \in S_i} \mathbf{z'} 
\quad \text{and} \quad
\mathbf{\sigma}_i^2 = \frac{1}{|S_i|} \sum_{\mathbf{z}' \in S_i} (\mathbf{z}' - \mathbf{c}_i)^2
\end{equation}

In the clustering-based approach to learning representations, the first step involves assigning each feature vector \( \mathbf{z} \) to a cluster centroid. This assignment is determined by:

\begin{equation}\label{eq:assignment}
\mathbf{c}_{i[\mathbf{z}]} = \underset{\mathbf{c}_l}{\text{argmin}} \ \|\mathbf{z} - \mathbf{c}_l\|
\end{equation}

\noindent where \( \mathbf{c}_l \) are the available centroids and \( \mathbf{c}_i \) is the normalized mean vector of the cluster to which \( \mathbf{z} \) is closest in terms of Euclidean distance. After this assignment, the learning objectives can be applied to refine the feature vectors with respect to their clusters.

\begin{figure*}[htbp]
    \centering
    \includegraphics[width=0.9\textwidth]{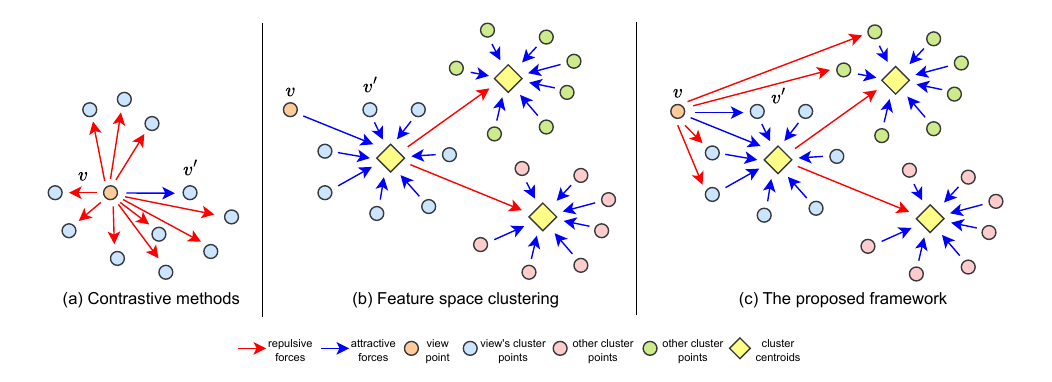}
    \caption{Visualization of the repulsive and attractive forces. (a) Illustrates the contrastive objective, emphasizing class separation (Section \ref{contrastivelearning}). (b) Applies feature space clustering to improve intra-class compactness (Section \ref{clustering}). (c) Demonstrates the CueCo framework, merging (a) and (b).}
    \label{fig:forces}
\end{figure*}

The first clustering objective, the centroid contrastive loss, aligns features with the centroid of their assigned cluster, ensuring they remain distinct from centroids of other clusters. This is achieved with an InfoNCE objective \cite{oord2019cpc} that measures the similarity between the features and their respective cluster centroids:

\begin{equation}\label{loss_centroid}
\mathcal{L}_\text{2}(\mathbf{z}, \mathbf{c}_{i[\mathbf{z}]}) = -\log \frac{\exp(\mathbf{z}^\top \cdot \mathbf{c}_{i[\mathbf{z}]} / \tau)}{\exp(\mathbf{z}^\top \cdot \mathbf{c}_{i[\mathbf{z}]} / \tau) + \sum\limits_{l=1}^{L} \exp(\mathbf{z}^\top \cdot \mathbf{c}_l / \tau)}
\end{equation}

\noindent where \( \mathbf{c}_{i[\mathbf{z}]} \) is the centroid to which \( \mathbf{z} \) has been assigned, \( \mathbf{c}_l \) are the centroids of other clusters, and \(\tau\) is a temperature parameter that adjusts the \(\ell_2\) normalized vectors (can be different from the temperature in contrastive loss). This loss encourages each feature vector \( \mathbf{z} \) to align closely with its own cluster centroid while remaining distinct from the centroids of other clusters, effectively refining the granularity of the learned representations.

The second objective, namely the variance loss, minimizes the squared Euclidean distance between features and their corresponding cluster centroid, scaled by the variance of the cluster:

\begin{equation}\label{eq:loss_var}
\mathcal{L}_\text{3}(\mathbf{z}, \mathbf{c}_{i[\mathbf{z}]}) = \frac{\left\| \mathbf{z} - \mathbf{c}_{i[\mathbf{z}]} \right\|^2}{2 \cdot \sigma_{i[\mathbf{z}]}^2 + \epsilon}
\end{equation}

\noindent where \( \sigma_i^2 \) is the variance of the cluster to which \( \mathbf{z} \) is assigned and \( \epsilon = 10^{-6} \) is a small constant to prevent division by zero. This objective ensures that not only do the features align with the centroids, but they also conform to the cluster's overall distribution, promoting uniformity within clusters and enhancing the robustness of the clustering.

\subsection{Final Objective} \label{objective}

The encoder minimizes a combination of the loss functions:

\begin{equation}\label{eq:loss}
\mathcal{L} = \lambda _1 \cdot \mathcal{L}_\text{1} + \lambda _2 \cdot \mathcal{L}_\text{2} + \lambda _3 \cdot \mathcal{L}_\text{3}
\end{equation}

\noindent where $\lambda_1, \lambda_2, \lambda_3$ are coefficients that balance the impact of each loss component on the representation learning. After training, only the convolutional encoder of \(f_\theta\) is as in \cite{he2020moco}, utilizing \(f_\xi\) during training to compute mean vectors and variances. The encoder \(f_\xi\) maintains a slow-moving average of \(f_\theta\), which stabilizes the feature representations over time and ensures a consistent interpretation of the feature space. This consistency is critical for maintaining reliable cluster centroids and standard deviations, thereby enhancing the overall robustness and accuracy of the clustering mechanism. Both the centroid contrastive and variance losses during the forward pass are computed using the outputs from \(f_\theta\), as the parameters need to be updated based on the current, directly observed representations, rather than the averaged historical data.

\subsection{Towards Online Feature Space Clustering} \label{momentumclustering}

Using an offline clustering algorithm like K-means is restrictive because it requires that the algorithm be applied at the end of each epoch to use the resulting cluster centroids and standard deviations for training objectives. Moreover, once established, the clustering remains unchanged for the duration of the epoch, rendering it static. This approach becomes particularly impractical for large datasets. To make the clustering algorithm even more tractable, we apply K-means in the queue to get the cluster centers and cluster standard deviations. Furthermore, cluster centroids need to reflect the most current instances to facilitate gradient propagation and must be updated in sync with \(f_\theta\) using a differentiable function \cite{pang2022smog}. To address these challenges, we introduce a momentum grouping scheme, similar to \cite{pang2022smog}. We initialize the cluster features \(\{\mathbf{c}_1, \mathbf{c}_2, \ldots\, \mathbf{c}_L\}\) randomly or via a method such as K-means and throughout the training process, we continuously update the centroids \(\mathbf{c}_i\) and standard deviations \(\sigma_i\) at each iteration as follows:

\begin{equation}\label{eq:updatecentroids}
\mathbf{c}_i \gets \beta_1 \cdot \mathbf{c}_i + (1 - \beta_1) \cdot \left( \frac{1}{|S_i|} \sum_{\mathbf{z} \in S_i} \mathbf{z}' \right)
\end{equation}

\begin{equation}\label{eq:updatestds}
\sigma_i^2 \gets \beta_2 \cdot \sigma_i^2 + (1 - \beta_2) \cdot \left( \frac{1}{|S_i|} \sum_{\mathbf{z} \in S_i} (\mathbf{z}' - \mathbf{\mu}_i )^2 \right)
\end{equation}

\noindent where $\mathbf{\mu}_i$ is the mean vector of features in cluster $i$ and \(\beta_1, \beta_2\) represent the momentum ratios. The momentum grouping method dynamically assigns each instance to the nearest centroid and updates the centroid and standard deviation features in a momentum-based manner. Consequently, the centroid and standard deviation features consistently represent the most recent visual characteristics of the instances.

\subsection{Intuition on Behavior} \label{intuition}

The dynamics within CueCo resemble the forces in electromagnetism, where embeddings are metaphorically ``charged'' to repel or attract each other based on the model's losses and architecture. The contrastive loss acts as a repulsive force, pushing apart embeddings of dissimilar instances to expand the feature space, akin to like charges repelling. In contrast, clustering-driven losses act as an attractive force, pulling together embeddings of similar instances to form cohesive clusters, similar to oppositely charged particles attracting. As illustrated in \Cref{fig:forces}, repulsive forces ensure wide separation among classes, while attractive forces (e.g., with $\lambda_2, \lambda_3 > 0$) pull embeddings into tightly-knit groups. The ultimate goal is to reach an equilibrium after $t$ training steps, where embeddings $\mathbf{z}_{t,Ti} = f(x_i; w_t)$ achieve optimal separation between dissimilar classes and cohesion within similar classes, resulting in robust, accurate representations.


\section{Experiments} \label{sec:experiments}

\begin{table}[!t]
\centering
\setlength{\tabcolsep}{1mm}
\caption{Linear top-1 and top-5 accuracies (\%) obtained through linear evaluation protocol on CIFAR-10, CIFAR-100, and ImageNet-100 datasets using ResNet-18 as the backbone network. Results adapted from \cite{sololearn}.}
\label{tab:results_cls}
\begin{tabular}{lcccccc}
\hline
\multirow{2}{*}{Method} & \multicolumn{2}{c}{CIFAR-10} & \multicolumn{2}{c}{CIFAR-100} & \multicolumn{2}{c}{ImageNet-100} \\
\cmidrule(lr){2-3} \cmidrule(lr){4-5} \cmidrule(lr){6-7}
 & Top-1 & Top-5 & Top-1 & Top-5 & Top-1 & Top-5 \\ \hline
BYOL \cite{grill2020byol} & 92.61 & 99.82 & 70.18 & 91.36 & 80.09 & 94.99 \\
DINO \cite{caron2021dino} & 89.19 & 99.31 & 66.38 & 90.18 & 74.84 & 92.92 \\
SimSiam \cite{chen2020simsiam} & 90.51 & 99.72 & 65.86 & 89.48 & 77.04 & 94.02 \\
MoCo-v2 \cite{chen2020mocov2} & 92.94 & 99.79 & 69.54 & 91.49 & 78.20 & 95.50 \\
MoCo-v3 \cite{chen2021mocov3} & 93.10 & 99.80 & 68.83 & 90.57 & 80.86 & 95.18 \\
ReSSL \cite{zheng2021ressl} & 90.63 & 99.62 & 65.83 & 89.51 & 76.59 & 94.41 \\
VICReg \cite{bardes2022vicreg} & 90.07 & 99.71 & 68.54 & 90.83 & 79.22 & 95.06 \\
SwAV \cite{caron2020swav} & 89.17 & 99.68 & 64.67 & 88.52 & 74.28 & 92.84 \\
SimCLR \cite{chen2020simclr} & 90.74 & 99.75 & 65.39 & 88.58 & 77.48 & 93.42 \\
BT \cite{zbontar2021barlowtwins} & 89.57 & 99.73 & 69.18 & 91.19 & 78.62 & 94.72 \\
CueCo (ours) & 91.40 & 99.79 & 68.56 & 91.05 & 78.65 & 94.03 \\
\hline
\end{tabular}
\end{table}

\begin{table*}[!htb]
\centering
\caption{Unsupervised image classification results on CIFAR-10 and CIFAR-100 datasets. We report standard clustering metrics including Normalized Mutual Information (NMI), Adjusted Mutual Information (AMI), Adjusted Rand Index (ARI), and clustering accuracy (ACC). All methods are reimplemented.}
\label{tab:results_cluster}
\begin{tabular}{lcccccccc}
\hline
\multirow{2}{*}{Method} & \multicolumn{4}{c}{CIFAR-10} & \multicolumn{4}{c}{CIFAR-100} \\
 \cmidrule(lr){2-5} \cmidrule(lr){6-9}
 & NMI & AMI & ARI & ACC & NMI & AMI & ARI & ACC \\ \hline
MoCo-v2 \cite{chen2020mocov2} (repr.) & 60.96 & 60.88 & 28.95 & 63.51 & 51.77 & 45.84 & 9.95 & 31.72 \\ 
SimCLR \cite{chen2020simclr} (repr.) & 69.03 & 68.98 & 53.14 & 74.50 & 50.75 & 44.60 & 11.15 & 32.17 \\ 
CueCo (ours) & \textbf{69.33} & \textbf{69.01} & \textbf{53.87} & \textbf{75.06} & \textbf{52.37} & \textbf{46.31} & \textbf{11.35} & \textbf{33.82} \\
\hline
\end{tabular}
\end{table*}

\begin{table*}[!htb]
\centering
\caption{Ablation studies on CIFAR-100 evaluating the impact of different loss terms. We report linear evaluation accuracies (top-1, top-5), k-nearest neighbor accuracies (20-NN, 100-NN), and clustering metrics (NMI, AMI, ARI, ACC). Checkmarks indicate which loss terms are active. All methods are reimplemented.}
\label{tab:objective_performance}
\begin{tabular}{ccccccccccc}
\hline
$\lambda_1$ & $\lambda_2$ & $\lambda_3$ & Top-1 & Top-5 & 20-NN & 100-NN & NMI & AMI & ARI & ACC \\ \hline
\checkmark &  &  & {67.9} & \underline{90.7} & {66.5} & {66.4} & 50.5 & 44.4 & 8.3 & 31.1 \\
\checkmark & \checkmark &  & 66.9 & 90.6 & 63.9 & 62.4 & \textbf{54.6} & \textbf{48.6} & \textbf{14.9} & \textbf{34.5} \\
\checkmark &  & \checkmark & \underline{68.4} & \underline{90.7} & \underline{66.7} & \underline{66.8} & {51.2} & {45.1} & 8.6 & {32.3} \\
\checkmark & \checkmark & \checkmark & \textbf{68.5} & \textbf{91.0} & \textbf{66.8} & \textbf{67.0} & \underline{52.3} & \underline{46.3} & \underline{11.3} & \underline{33.8} \\
\hline
\end{tabular}
\end{table*}

\subsection{Implementation Details} \label{implemntationdetails}

We evaluate CueCo on CIFAR-10, CIFAR-100 \cite{krizhevsky2009cifar}, and ImageNet-100 \cite{deng2009imagenet}. Each input image is transformed twice to generate two different views using the same augmentation as used in \cite{grill2020byol}. Our encoder $f_\theta$ includes a ResNet-18 backbone (adapted by replacing the $7 \times 7$ convolution (\verb|conv1|) with a $3 \times 3$ convolution and removing the max pooling layer), a 2-layer MLP projection head, and a prediction head \cite{chen2021mocov3}. The encoder $f_\xi$ shares the backbone and projection head but excludes the prediction head and is updated as a moving average of $f_\theta$ \cite{he2020moco, grill2020byol}. Both MLPs have hidden layers of 4096-d with ReLU \cite{nair2010relu}, 256-d output layers without ReLU, and batch normalization \cite{ioffe2015batch}. For pretraining we use the SGD optimizer \cite{ruder2017sgd} with a base learning rate of $0.3$ (decreased to 0 with cosine schedule), momentum of $0.9$, and weight decay of $10^{-4}$. Clustering is frozen for the first 313 iterations to prevent instability, and cluster features are reset every 1,000 iterations to avoid collapse and maintain balanced cluster distributions (following \cite{caron2020swav,pang2022smog}). We train CIFAR-10/1000 for 1000 epochs and ImageNet-100 for 400 epochs. We use a FIFO queue of size 16,384 from which we calculate the cluster centroids and variances (but updated via momentum to reduce computational overhead), and the momentum coefficient for the moving average encoder starts at 0.996 and increases to 1.0 using a cosine schedule. All temperatures are set to $\tau = 0.2$, and loss weights are $\lambda_1 = 1$, $\lambda_2 = 0.1$, and $\lambda_3 = 0.01$. 

\subsection{Linear Evaluation} \label{initeval}

We evaluate CueCo's representations by training a linear classifier on top of the frozen features from the ResNet-18 encoder. The linear classifier is trained for 100 epochs using a learning rate of 3.0 with a cosine learning rate scheduler. Training minimizes the cross-entropy loss with an SGD optimizer, momentum of 0.9, and weight decay of \(1 \times 10^{-6}\), using a batch size of 256. We report top-1 and top-5 accuracies as percentages on the test set in \Cref{tab:results_cls}, comparing our results to previous state-of-the-art self-supervised methods from \cite{sololearn}. Our method performs competitively with state-of-the-art self-supervised approaches, while not being heavily tuned.

\subsection{Unsupervised Image Classification}

We evaluate our approach on the task of unsupervised image classification for CIFAR-10/100. We report the standard clustering metrics: Normalized Mutual Information (NMI), Adjusted Normalized Mutual Information (AMI), Adjusted Rand-Index (ARI), and Clustering Accuracy (ACC), as in \cite{amrani2022selfclassifier}. We compare our approach with reproductions of MoCo-v2 \cite{chen2020mocov2} and SimCLR \cite{chen2020simclr}, using the same hyperparameter settings for a fair comparison. As shown in \Cref{tab:results_cluster}, our method, CueCo, consistently outperforms the compared methods across all metrics on both CIFAR-10 and CIFAR-100.

\subsection{Ablation Study} \label{ablation}

We investigate the importance of each objective in our method, focusing on contrastive, centroid contrastive, and variance reduction losses. \Cref{tab:objective_performance} presents results on CIFAR-100 using different loss combinations. Contrastive loss alone provides a strong baseline, while adding centroid contrastive loss improves clustering metrics. Incorporating variance reduction loss further enhances linear evaluation and unsupervised classification. These findings highlight the effectiveness of our multi-objective approach in producing versatile and discriminative representations by balancing instance-level distinctiveness and class-level coherence.


\section{Conclusion}

CueCo advances unsupervised visual representation learning by integrating contrastive learning with momentum clustering, creating a 	``push-pull'' dynamic that simultaneously enhances inter-class separation and intra-class cohesion. The framework demonstrates competitive performance on benchmark datasets while particularly excelling in unsupervised image classification metrics. By balancing these complementary forces, CueCo establishes a promising direction for self-supervised learning that produces more discriminative representations with improved class structure without requiring labeled data.






\bibliographystyle{IEEEbib}
\bibliography{refs}

\end{document}